\documentclass[letterpaper, 10 pt, conference]{ieeeconf}  

\IEEEoverridecommandlockouts                              

\title{\LARGE \bf
Revisiting Matching Response and Swept Feature Volumes for Wide-baseline Omnidirectional Stereo}

\author{Seungjin Jeon$^{1*}$ Jongwoo Lim$^{1,2}$ and Changhee Won$^{1*\dagger}$
\thanks{$^*$Authors contributed equally.}
\thanks{$^\dagger$Corresponding author.}
\thanks{$^{1}$UVify Corporate Affiliated Research Institute, Republic of Korea.
         {\tt\small \{seungjin.jeon, changhee.won\}@uvify.com}}
\thanks{$^{2}$Seoul National University, Seoul, Republic of Korea.
         {\tt\small \{jongwoo.lim\}@snu.ac.kr}}
}

\usepackage{graphicx}
\usepackage{amssymb}
\usepackage{amsmath}
\usepackage{booktabs,multirow,array}
\usepackage{adjustbox}
\usepackage{longtable}
\usepackage{gensymb}
\usepackage[font=small,labelfont=bf,singlelinecheck=false]{caption}
\usepackage[subrefformat=parens]{subcaption}
\usepackage{bm}
\let\labelindent\relax
\usepackage{enumitem}
\usepackage{subcaption}
\usepackage{color}
\usepackage{siunitx}
\usepackage{threeparttable}
\usepackage{tabularx}
\usepackage[top=19.5mm, bottom=19.5mm, left=19.2mm, right=19.2mm]{geometry}

\makeatletter
\let\NAT@parse\undefined
\makeatother

\usepackage{cite}
\usepackage{hyperref}
\hypersetup{colorlinks,linktocpage=true}

\newcommand{\tb}[1]{\textbf{#1}}
\newcommand{\mb}[1]{\mathbf{#1}}
\newcommand{\etal}[0]{\textit{et al.}}

\newcommand{\subsec}[1]{\vspace{4pt}{\setlength{\parindent}{0pt}\tb{#1}~}}

\newcommand{\Figure}[1]{Fig.~\ref{#1}}
\newcommand{\Table}[1]{Tbl.~\ref{#1}}
\newcommand{\Equation}[1]{Eq.~\ref{#1}}

\begin{document}

\maketitle
\thispagestyle{empty}
\pagestyle{empty}

\begin{abstract}
In this paper, we propose a training strategy for confidence estimation in omnidirectional stereo, targeting the ambiguous matches that frequently occur in wide-baseline setups. Reinterpreting the matching responses produced by the 3D encoder–decoder block, we show that their expectation values provide intrinsic confidence signals. Building on this, our method directly penalizes ambiguous responses without auxiliary heads, multi-pass inference, or additional modules, resulting in more efficient and generalized predictions. Beyond confidence, we introduce swept feature volume resampling, where response features produced by 3D CNNs are resampled using regressed positive matching indices and then processed by 2D CNNs to predict meta-information such as surface normals. This joint learning introduces auxiliary geometric regularization and improves depth coherence by leveraging additional contextual cues during response aggregation stage. Experimental results demonstrate that our approach enhances both confidence estimation and surface normal prediction while maintaining deployment practicality for autonomous mobility applications.
\end{abstract}

\section{INTRODUCTION}

Omnidirectional perception has become an important component in various computer vision applications, particularly in robotics, autonomous driving, and aerial navigation, where accurate depth estimation critically affects system performance and safety. 
Although LiDAR sensors can be employed, recent advances in deep neural networks (DNNs) have greatly improved stereo matching~\cite{kendall2017end,chang2018pyramid}, making multi-camera omnidirectional stereo approaches increasingly attractive. 
Existing omnidirectional stereo methods typically involve configurations with multiple cameras~\cite{won2020end,won2019omnimvs,won2019sweepnet,deng2025omnistereo,jiang2024romnistereo,xie2023omnividar} or stereo setups using 360$\degree$ cameras~\cite{wang2020360sd,zayene2025helvipad,li2022mode}. 
More recently, approaches that employ wide-baseline cameras with global spherical sweeping have been proposed~\cite{won2020end,won2019omnimvs,won2019sweepnet,li2022mode,deng2025omnistereo,jiang2024romnistereo}.

\begin{figure}[t]
    \centering
    
    \includegraphics[width=\linewidth]{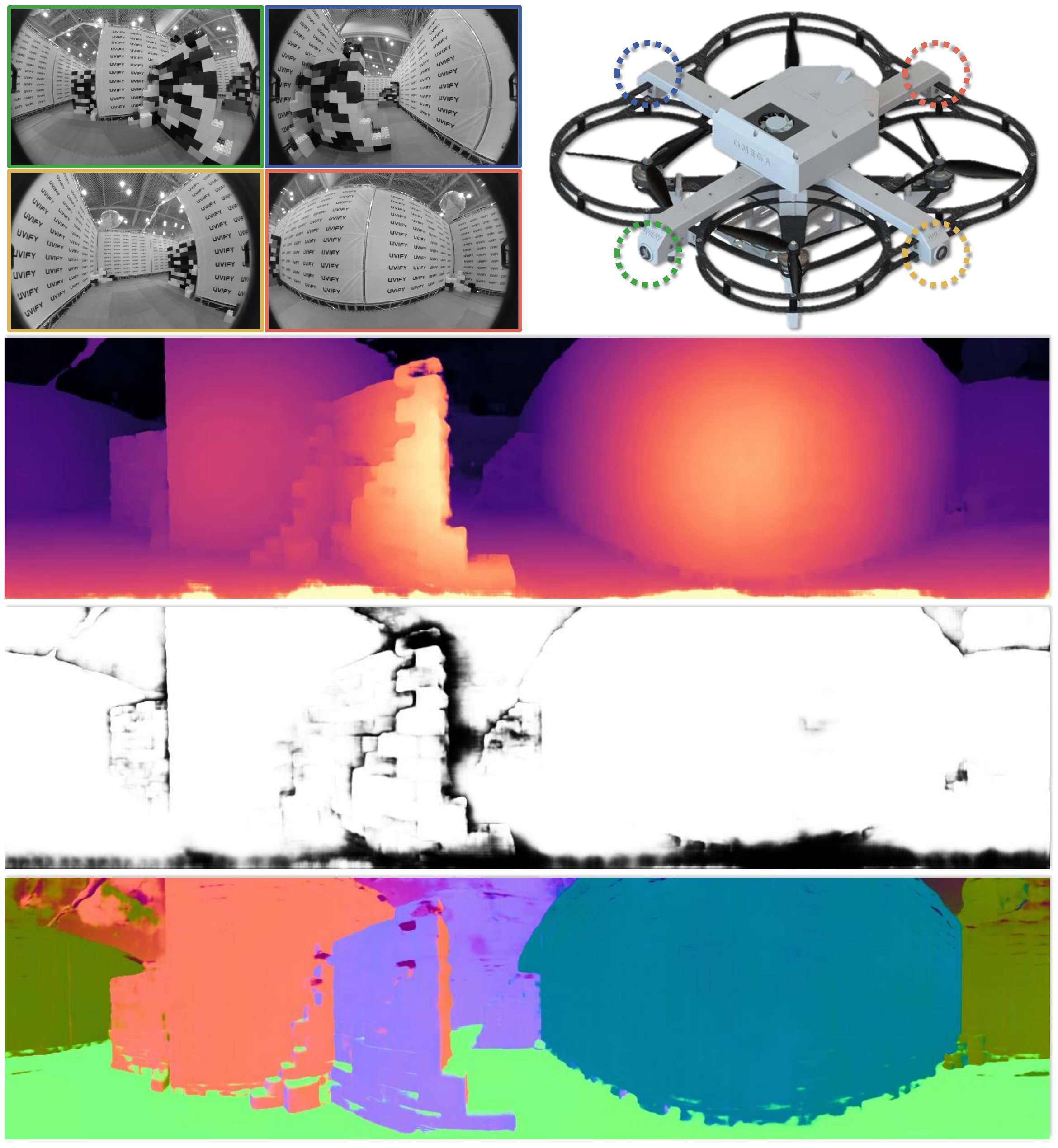}
    \caption{
    From top: four wide field-of-view (FoV) fisheye input images and our multi-camera equipped drone, followed by inverse-depth prediction, confidence prediction, and surface normal prediction from the proposed method. 
    }

    \label{fig:introduction}
\end{figure}

In wide-baseline setups with global sweeping, occlusions without positive (valid) matches often lead to erroneous depth estimation results. 
For practical deployment on mobile platforms such as robots and drones, it is therefore essential to detect and handle such ambiguous matches using a confidence measure of the matching estimates. 
Various confidence estimation methods have been proposed for stereo matching.
Poggi and Mattoccia~\cite{poggi2016learning}, for example, feed the output disparity into separate 2D convolutional neural networks (CNNs) to predict confidence.
Similarly, contextual features from input images~\cite{tosi2018beyond}, cost volumes~\cite{kim2019laf}, or iteratively computed disparity profile~\cite{lee2024modeling} have been used as inputs to dedicated confidence estimation networks.
Auxiliary heads within stereo matching networks have also been employed to predict confidence~\cite{zhang2020adaptive,chen2023learning}.
However, such approaches typically incur higher inference time and architectural complexity, since they rely on additional network modules that go beyond the core stereo matching framework, limiting their practicality for real-time deployment.
Without additional modules, Won~\etal~\cite{won2020end} use entropy as an uncertainty measure and guide the network to produce more confident (low-entropy) matching results by applying a hinge loss to the output entropy.
Although the entropy measure from normalized matching probability volumes effectively detects textureless or repetitive regions that yield flat matching responses, in wide-baseline setups with global sweeping, occlusions without positive (valid) matches cause the entropy to behave randomly, revealing a key limitation. 
In this paper, we propose a training strategy for confidence estimation to detect ambiguity caused by wide-baseline setups.
We reinterpret the matching response volume and introduce an effective network-guided learning strategy based on this reinterpretation.

In addition to depth estimation, recent studies have explored joint learning of meta-information with stereo matching to enhance both the accuracy and consistency of depth. 
Pixel-wise semantic segmentation~\cite{dovesi2020real,chen2020sgnet,zhou2025all} and surface normal prediction~\cite{kusupati2020normal,liu2022digging,qi2018geonet} can serve as cues to determine whether two pixels (or rays) belong to the same or different surfaces/objects, thereby constraining correspondence search and boosting matching performance. 
However, in global sweeping pipelines~\cite{won2020end,won2019omnimvs,won2019sweepnet}, the input and output views have different optical centers, which hinders the direct application of such meta-information learning. 
To address this limitation, we propose \emph{Swept Feature Volume Resampling}, a mechanism that enables meta-information learning even under global sweeping with mismatched optical centers. 
As a concrete instantiation, we demonstrate that incorporating surface normal estimation within \emph{Swept Feature Volume Resampling} yields consistent improvements in both depth accuracy and consistency.

\Figure{fig:introduction} illustrates our multi-camera system for omnidirectional multi-view stereo matching, along with example results of inverse-depth estimation, confidence prediction, and surface normal prediction. 
The main contributions of this work are summarized as follows:
\begin{itemize} 
\item
We introduce a reinterpretation of the matching response volume for confidence estimation and propose modified loss functions that guide the network to properly control the matching responses.
Our confidence measure effectively penalizes ambiguous regions in wide-baseline setups without requiring auxiliary heads or additional network modules.
\item
We present a resampling strategy for swept feature volumes that enables joint learning in the global sweeping approach for wide-baseline multi-view stereo.
The resampled feature volumes are processed by separate networks to learn meta-information such as surface normals, thereby embedding additional contextual cues into the matching response computation stage.
Experiments show that the proposed surface normal prediction head provides auxiliary supervision that encourages geometric consistency during response aggregation.
\item
Extensive experiments on real and synthetic datasets demonstrate that our method outperforms previous confidence estimation approaches by more effectively filtering out ambiguous regions.
The proposed framework maintains efficiency suitable for practical deployment in robotics applications.
\end{itemize}

\section{Related Work}

\subsection{Omnidirectional Depth Estimation}
Omnidirectional depth estimation has advanced rapidly in recent years~\cite{won2019omnimvs,won2019sweepnet,won2020end,xie2023omnividar,komatsu2020360,jiang2024romnistereo,deng2025omnistereo,li2022mode}. 
Representative approaches based on global sweeping extend plane-sweep~\cite{collins1996space} to the spherical domain, enabling effective fusion of multi-fisheye views. 
In particular, OmniMVS and SweepNet~\cite{won2019omnimvs,won2019sweepnet,won2020end} leverage spherical sweeping to construct response volume and have demonstrated strong performance for omnidirectional scenes. 
Building on cost-volume reasoning, Romnistereo~\cite{jiang2024romnistereo} incorporates recurrent GRU aggregation—originally popularized in stereo matching~\cite{lipson2021raft}—to refine matching and improve robustness.
CrownConv360~\cite{komatsu2020360} further enhances spherical matching by adopting an icosahedron-based discretization of the sphere, yielding more uniform angular sampling and better geometric fidelity. 
In parallel, Omnividar~\cite{xie2023omnividar} proposes a fisheye-aware camera model that reduces projection distortion and validates performance by reusing stereo-matching backbones. 
Diverging from global sweeping, MODE~\cite{li2022mode} and OmniStereo~\cite{deng2025omnistereo} employ the Cassini projection to recast omnidirectional inputs into stereo-friendly coordinates, allowing the use of standard stereo pipelines without spherical cost construction.

\begin{figure}[t!]
    \centering
    \includegraphics[keepaspectratio=true,width=\linewidth]{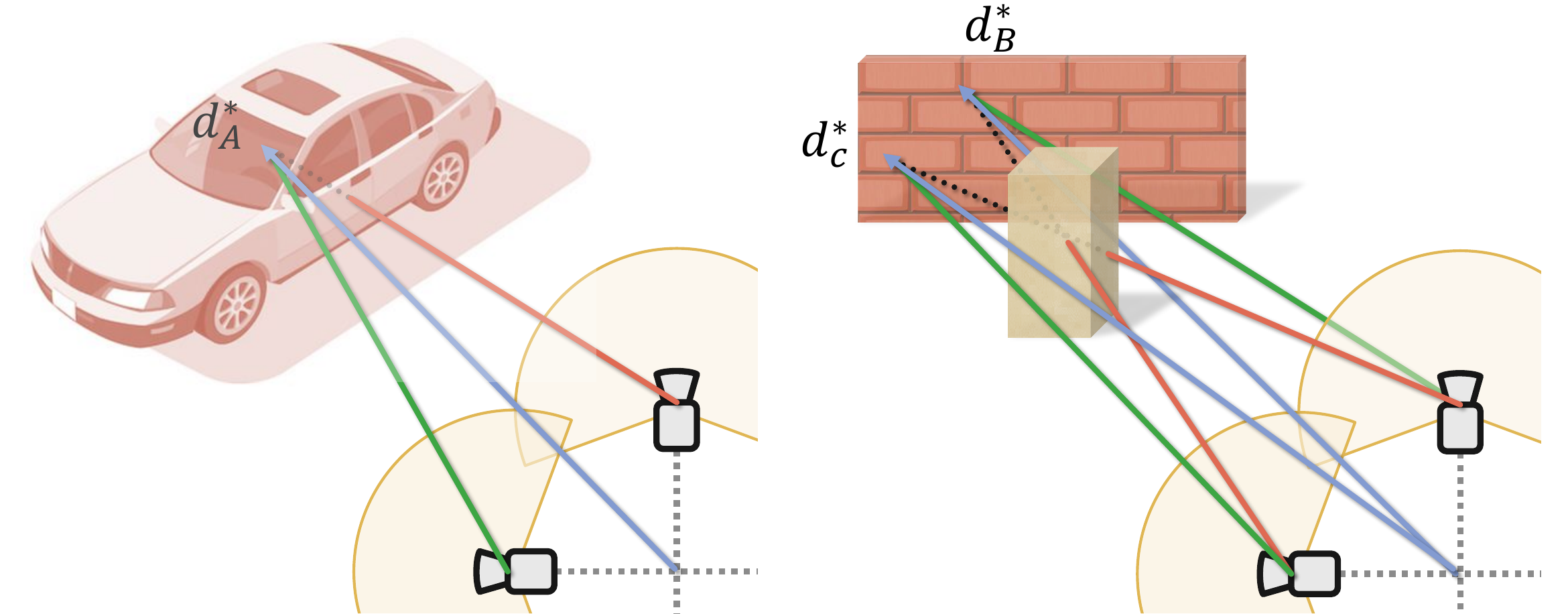}
    \caption{
    \tb{Examples of ambiguous matching caused by occlusions.} 
    In wide-baseline setups with the global sweeping approach, occlusion cases frequently occur: different cameras may observe different sides of an object ($A$), or backgrounds may be occluded by foreground objects ($B$ and $C$). 
    In such cases, the normalized matching probabilities become random since no positive match exists.
    }
    \label{fig:motive_confidence}
\end{figure}

Despite these advances, wide-baseline camera layout has hard matching cases (e.g., occlusions and large viewpoint changes), as illustrated in Fig.~\ref{fig:motive_confidence}. 
We propose a framework that estimates per-pixel confidence to mitigate ambiguity and occlusion-induced mismatches. 
Moreover, because the optical centers of the input and the output are different, many pipelines under-utilize meta-information and contextual image features. 
To address these limitations, we propose \emph{Swept Feature Volume Resampling} that enables the use of contextual features of the inputs.

\begin{figure*}[t]
\centering
    \includegraphics[keepaspectratio=true,width=\textwidth]{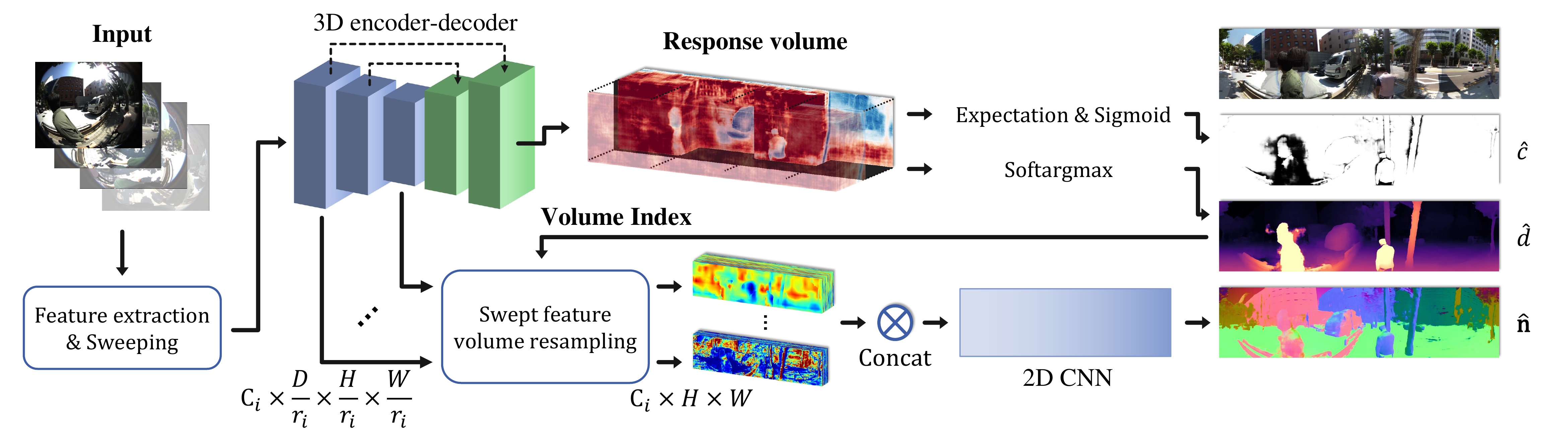}
    \caption{
    Input images are processed by 2D CNNs and swept onto global spheres, after which the 3D encoder–decoder block computes matching responses.
    Inverse-depth prediction is obtained by applying the softargmax operation to the matching response volume.
    We compute the expectation value of the output response volume and convert it into a confidence measure using a sigmoid function. 
    Meanwhile, using the inverse-depth index, we resample intermediate feature maps from the 3D encoders, concatenate them, and feed them into a 2D CNN to obtain surface normal predictions.
    }
    \label{fig:workflow}
\end{figure*}

\subsection{Confidence Estimation in Stereo Matching}
In stereo matching, confidence estimation has evolved from hand-crafted measures to learning-based approaches; over the past decade, deep neural network methods have delivered substantial performance gains~\cite{poggi2021confidence}. 
CCNN~\cite{poggi2016learning} estimates the confidence of disparities produced by a stereo matcher using a 2D CNNs. Building on this line of work, LAF-Net~\cite{kim2019laf} further advanced this direction by adaptively fusing multiple local confidence cues. 
Additional gains have been reported by broadening the input modality: from single-input (disparity-only)~\cite{poggi2016learning} designs to dual-input (e.g., disparity + image)~\cite{fu2017stereo} and triple-input (image, cost volume, disparity)~\cite{kim2019laf} architectures. More recently, Lee~\etal~\cite{lee2024modeling} leveraged plane-sweep to boost existing confidence networks. 
These methods, Poggi~\etal~\cite{poggi2016learning} tend to be highly dataset-dependent, and dual/triple-input confidence networks~\cite{kim2019laf,fu2017stereo} are require extra confidence heads or auxiliary modules. 

In contrast, our method uses neither an additional confidence head nor an auxiliary network, yet it effectively handles ambiguous matching regions.

\subsection{Meta-information Learning in Stereo Matching}
Recent stereo-matching research increasingly augments correspondence with auxiliary image cues—semantic segmentation~\cite{dovesi2020real,chen2020sgnet}, predicted surface normals~\cite{kusupati2020normal,liu2022digging,qi2018geonet,bartolomei2025stereo}, edges and instance masks, and, more recently, priors distilled from foundation models~\cite{wen2025foundationstereo,zhou2025all,cheng2025monster}. 
These cues act as scene-structure regularizers: they sharpen object boundaries, enforce piecewise planarity, and stabilize depth in textureless regions. 
These pipelines that exploit such cues are implicitly camera-centric: they build plane-sweep or cost volumes anchored at each input camera and assume that the auxiliary predictions and the target depth are defined in the same coordinate frame and share the same optical center. 
This assumption breaks down in global-sweeping models~\cite{won2020end,won2019omnimvs,won2019sweepnet} that parameterize the sweep around a rig-level (virtual) center. 
In this setting, the input and output views have different optical centers, which hinders the direct use of meta-information: per-view cues predicted in image coordinates do not align with a rig-centered sweep, leading to optical center inconsistencies. 

To resolve this optical-center mismatch, we introduce \emph{Swept Feature Volume Resampling}. 
It enables the use of contextual features from the input images even when the optical centers differ. 
Although we demonstrate its use for surface normals, the mechanism is generic and can also accommodate other per-view meta-information, such as semantic cues or foundation-model priors.


\section{Method}

In this section, we describe confidence estimation and swept feature resampling methods for wide-baseline omnidirectional stereo.
For multi-view stereo we mainly adopt the spherical sweeping procedure, network architectures, and training loss functions proposed in OmniMVS~\cite{won2020end}. The input fisheye images are first processed by 2D CNNs and then projected into spherical feature maps.
The 3D encoder-decoder block subsequently computes and aggregates the matching responses, and the final inverse-depth indices are obtained through the softargmax~\cite{kendall2017end}.
\Figure{fig:workflow} illustrates the overall procedure of our proposed methods.

\subsection{Confidence Estimation}

\begin{figure}[t!]
\centering
\resizebox{\linewidth}{!}{%
    \centering
    \includegraphics[width=\linewidth]{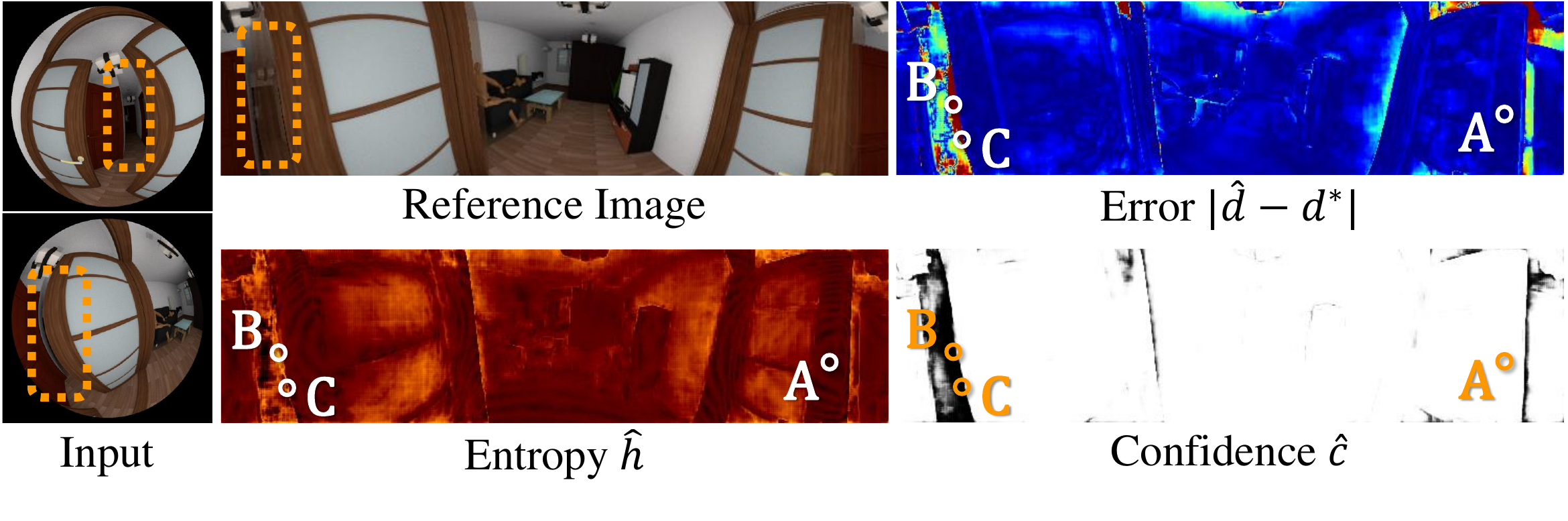}
    \label{fig:prediction_example}
}

\subfloat[]{
    \includegraphics[width=.5\linewidth]{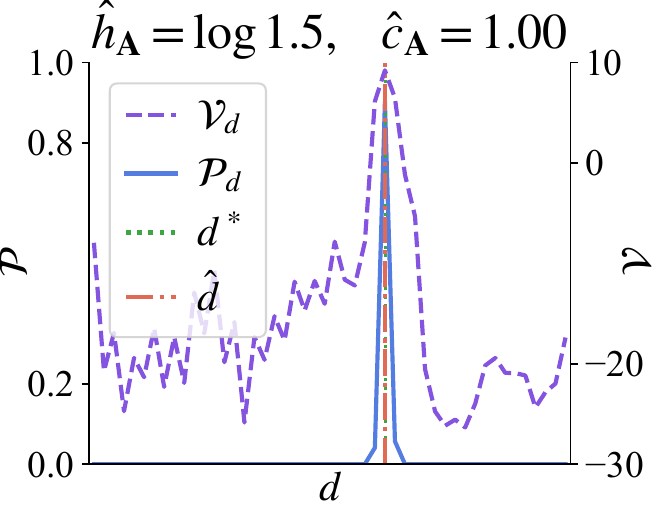}
    \label{fig:match_both_detect}
    }
    
\resizebox{\linewidth}{!}{
    \subfloat[]{
    \includegraphics[width=.5\linewidth]{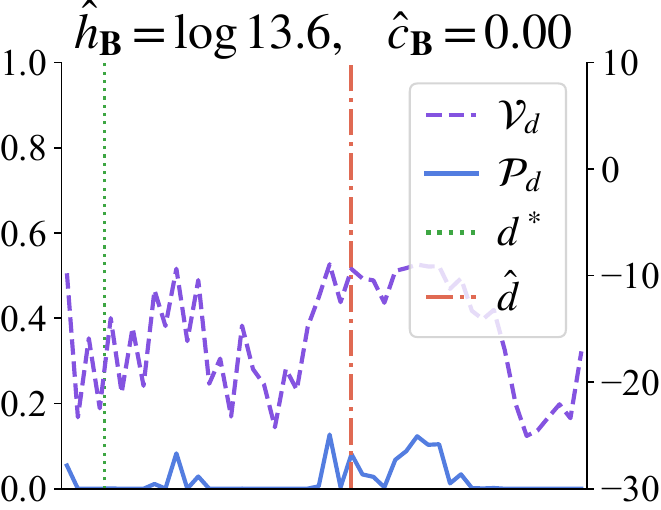}
    \label{fig:match_ent_fail}
    }
    \subfloat[]{
    \includegraphics[width=.5\linewidth]{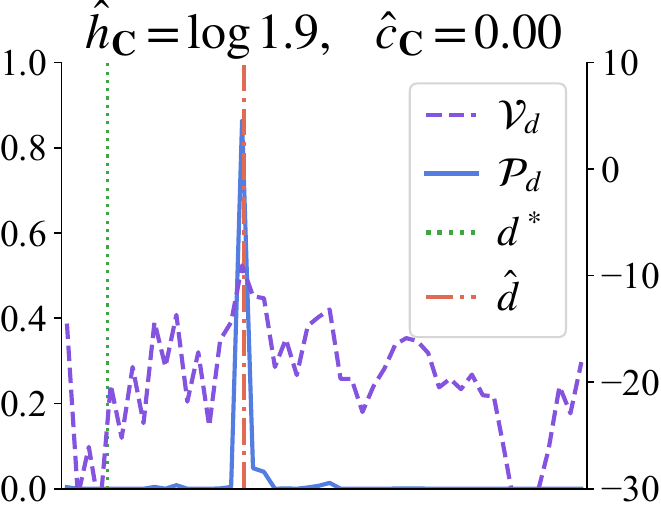}
    \label{fig:match_C}
    }
}

\caption{
(a) Example of a confident prediction at pixel $\mb{A}$. 
(b, c) Examples of erroneous predictions at pixels $\mb{B}$ and $\mb{C}$ in occluded regions (highlighted by the orange rectangles). 
The normalized probability and entropy (\Equation{eq:p_entropy}) produce inconsistent results on $\mb{B}$ and $\mb{C}$, whereas our confidence measure (\Equation{eq:confidence}) outputs low values, 
demonstrating its ability to reliably filter out ambiguous regions.
}
\label{fig:matching_example}
\end{figure}

To recognize and penalize ambiguous matches in reflective or occluded regions, separate networks can be employed to estimate matching confidence~\cite{xiao2018confidence,poggi2016learning,kim2019laf,tosi2018beyond}. However, this approach incurs additional computational cost and increases inference time, making it less practical for real applications.
Without auxiliary heads, OmniMVS~\cite{won2020end} instead uses the entropy of the matching probability volume as an uncertainty measure guiding the network to produce fewer positive responses via the entropy boundary loss.
Nevertheless, in ultra-wide baseline setups, occluded cases without any valid matching response along the given hypotheses are frequently observed, as illustrated in~\Figure{fig:motive_confidence}, and \Figure{fig:match_ent_fail} shows that the entropy often fails to indicate such cases.

\subsec{Reinterpretation of Matching Responses}
Instead of relying on entropy, we directly interpret the magnitude of the matching responses produced by the 3D CNNs as a positiveness score.
The matching response volume $\mathcal{V}$ of size $D \times H \times W$ 
is normalized along the inverse-depth dimension using the softmax function 
\begin{equation}
\label{eq:softmax}
\mathcal{P}_d=\frac{e^{\mathcal{V}_d}}{\sum_{i} e^{\mathcal{V}_i}}.
\end{equation}
where $D$ denotes the number of inverse-depth indices, $H$ and $W$ represent the height and width of the output equirectangular inverse-depth maps, respectively.
The output inverse-depth index $\hat{d}$ is computed by the softargmax~\cite{kendall2017end} as
\begin{equation}
\label{eq:softargmax}
\hat{d}= \sum_{d=0}^{D-1} (d  \mathcal{P}_d).
\end{equation}
Similar to inverse-depth index regression, we compute the expectation value of the matching response $\hat{v}$ and convert it to confidence $\hat{c}$ as
\begin{equation}
\hat{v} = \sum_{d=0}^{D-1} (\mathcal{V}_d \mathcal{P}_d),
\end{equation}
\begin{equation}
\hat{c} = \sigma(\lambda_v \hat{v})
\label{eq:confidence}
\end{equation}
where $\lambda_v$ is a scaling parameter, and $\sigma(\cdot)$ denotes the sigmoid function which maps the representative response value into the range $[0,1]$; a score close to $1$ indicates high confidence.
To preserve negative values in the feature volumes in the response computation stage, we replace the activation layers in the 3D CNNs with ELU~\cite{clevert2015fast}.
\[
    \operatorname{ELU}(x)=
        \begin{cases}
            x & x > 0\\ 
            e^x - 1 & x \le 0
        \end{cases}
\]

\subsec{Guiding Networks}
We adopt a regression loss between the predicted inverse-depth index $\hat{d}$ and the ground truth ${d^*}$ using the smooth $L_1$ loss $\mathcal{L}_{\text{smooth-L1}}$, which is robust to outliers.
Since the response volume is normalized during regression (\Equation{eq:softmax}), the networks cannot effectively control the magnitude of responses.
To enforce low confidence in ambiguous matches, we modify the regression loss as follows:
\begin{equation}
\mathcal{L}_{\text{conf}}=(1 + \lambda_c \hat{c}) \mathcal{L}_{\text{smooth-L1}}(\hat{d},d^*)
-\lambda_c\operatorname{log}\hat{c},
\label{eq:loss_conf}
\end{equation}
where the index error is down-weighted in low-confidence regions, and $\lambda_c$ is a scaling parameter that adjusts the influence of the confidence score $\hat{c}$ on the regression loss.
At the same time, we encourage the confidence score to remain close to $1$ (high confidence), thereby preventing it from collapsing toward low values.




\subsection{Swept Feature Volume Resampling}

Meta information such as instance segmentation or surface normals has been exploited through joint learning with stereo matching to improve depth quality and consistency~\cite{dovesi2020real,chen2020sgnet,zhou2025all,wen2025foundationstereo,kusupati2020normal,liu2022digging,qi2018geonet}. 
To incorporate contextual cues into stereo matching, auxiliary features are often extracted from the reference view and fused into the matching blocks. 
However, in the global sweeping approach, where the sweeping center is not aligned with any camera, features from a single view cannot be directly fused into the swept features through simple operations such as concatenation or addition. 
To address this, we resample the feature volumes swept and fused from multiple views using the predicted inverse-depth index $\hat{d}$.

\subsec{4D Feature Volume Resampling}
Considering the 4D feature volumes $\mathcal{F}_i$ of size $C_i \times \frac{D}{r_i} \times \frac{H}{r_i} \times \frac{W}{r_i}$ output by the $i$-th 3D convolutional layer, we obtain the resampled volume $\bar{\mathcal{F}}_i$ of size $C_i \times H \times W$:
\begin{equation}
\bar{\mathcal{F}}_i(c, y, x) = \mathcal{F}_i\!\left(c, \tfrac{\hat{d}(y,x)}{r_i}, \tfrac{y}{r_i}, \tfrac{x}{r_i}\right),
\end{equation}
where $r_i$ denotes the reduction factor. 
We use trilinear interpolation to perform the feature volume resampling.
Since the resampled features are aligned with the sweeping center, they can be passed to auxiliary 2D networks to produce meta information.
To keep the resampling and auxiliary processing optional at inference time, we do not fuse features from the 2D networks into the matching blocks.
This allows the networks to exploit contextual cues during response computation and aggregation in the training stage, while avoiding additional overhead during deployment.

\subsec{Surface Normal Prediction}
In practice, we demonstrate surface normal prediction as an auxiliary task to enforce geometric consistency in the matching blocks. 
We design lightweight 2D networks with a few convolutional layers, which take the resampled feature volumes as input after concatenating them along the channel dimension. 
The 2D CNNs output normalized normal vectors $\hat{\mb{n}}$ of size $H \times W \times 3$. 
Using the ground-truth inverse-depth index $d^*$, we compute the corresponding ground-truth normal vectors $\mb{n}^*$.
The normal prediction loss is defined as the angular difference between the predicted and ground-truth normal vectors:
\begin{equation}
    \mathcal{L}_{\text{normal}} = \operatorname{arctan2}\!\left(\left\|\hat{\mb{n}} \times \mb{n}^*\right\|_2, \hat{\mb{n}} \cdot \mb{n}^* \right).
    \label{eq:loss_normal}
\end{equation}
The overall training loss terms is defined as follows:
\begin{equation}
    \mathcal{L} = \mathcal{L}_{\text{conf}} + 
    \lambda_n\mathcal{L}_{\text{normal}},
\end{equation}
where $\lambda_n$ is the weighting parameter of the normal prediction loss.


\begin{table*}[t!]
\setlength{\tabcolsep}{4pt}         
\renewcommand{\arraystretch}{1.1}     

\caption{\tb{Quantitative comparison of confidence measures.} 
Each network is evaluated both with and without fine-tuning. 
The metrics are defined in \Equation{eq:auc} and \Equation{eq:opt_auc}. 
AUC values are averaged over all test frames. 
The lowest AUC value is highlighted, and \textquotesingle$^*$\textquotesingle~denotes the second-best result.
\textquotesingle$^\dagger$\textquotesingle~denotes our proposed confidence measures.
%
}

\resizebox{\textwidth}{!}{

\begin{tabular}{l|r|rrrrr|rcrrrrr|r}
\noalign{\global\arrayrulewidth=0.8pt}
\cline{1-8}\cline{10-15}
\noalign{\global\arrayrulewidth=0.4pt}
Dataset & $C$ & 
\multicolumn{1}{c}{$^\dagger\gamma_{1, 10^{\tb{-6}}}$} &
\multicolumn{1}{c}{$^\dagger\gamma_{1, 10^{\tb{-3}}}$} &
\multicolumn{1}{c}{$^\dagger\hat{c}$} &
\multicolumn{1}{c}{$\hat{p}$~\cite{deng2025omnistereo, li2022mode}} &
\multicolumn{1}{c|}{$\hat{h}$~\cite{won2020end}} & \multicolumn{1}{c}
{OptAUC} &

\multirow{13}{*}{~\rotatebox[origin=c]{90}{Fine-tuned}} &

\multicolumn{1}{c}{$^\dagger\gamma_{1, 10^{\tb{-6}}}$} &
\multicolumn{1}{c}{$^\dagger\gamma_{1, 10^{\tb{-3}}}$} &
\multicolumn{1}{c}{$^\dagger\hat{c}$} &
\multicolumn{1}{c}{$\hat{p}$} &
\multicolumn{1}{c|}{$\hat{h}$} & \multicolumn{1}{c}{OptAUC} \rule{0pt}{3ex} \\

\cline{1-8}\cline{10-15}
\multirow{3}{*}{Sunny}
& 16     & \tb{0.672} & 0.727 & $^*$0.680 & 2.435 & 2.731 & 0.108 
& &\tb{0.248} & 0.315 & $^*$0.249 &0.791 &0.777 & 0.025 \\

&  8    & \tb{0.823} & 0.850 & $^*$0.837 & 2.581 & 2.901 & 0.141 
& &\tb{0.417} & 0.462 & $^*$0.421 & 0.981 & 0.960 & 0.049 \\

&  4     & \tb{1.037} & $^*$1.038 & 1.046 & 2.641 & 3.052 & 0.168 
&  &\tb{0.474} &0.486 & $^*$0.476 &1.076 &0.974 &0.067 \\

\cline{1-8}\cline{10-15}
\multirow{3}{*}{Cloudy}
& 16   & \tb{0.689} & 0.757 & $^*$0.698 & 2.567 & 2.934 & 0.096 
& &\tb{0.194} &0.271 & $^*$0.195 &0.650 &0.661 &0.016  \\
&  8     & \tb{0.749} & 0.788 & $^*$0.762 & 2.736 & 3.106 & 0.124 
& &\tb{0.329} &0.367 & $^*$0.335 &0.807 &0.840 &0.030  \\
&  4   & \tb{0.914} & $^*$0.917 & 0.919 & 2.592 & 2.944 & 0.144 
& &\tb{0.402} &0.420 & $^*$0.403 &0.944 &0.845 &0.049 \\

\cline{1-8}\cline{10-15}
\multirow{3}{*}{Sunset}
& 16 & \tb{0.599} & 0.629 & $^*$0.609 & 2.144 & 2.346 & 0.097 
& &\tb{0.260} &0.330 & $^*$0.261 &0.774 &0.760 &0.025 \\
&  8    & \tb{0.695} & 0.720 & $^*$0.708 & 1.996 & 2.161 & 0.120 
& &\tb{0.419} &0.627 & $^*$0.424 &0.943 &0.948 &0.050 \\
&  4    & \tb{0.912} & $^*$0.913 & 0.919 & 2.210 & 2.482 & 0.151 
& &\tb{0.499} &0.509 & $^*$0.500 &1.057 &0.982 &0.070 \\

\cline{1-8}\cline{10-15}
\multirow{3}{*}{OmniHouse}
& 16  & \tb{2.378} & 2.505 & $^*$2.427 & 3.808 & 4.097 & 1.150 
& & $^*$0.776 & \tb{0.773} & 0.846 & 1.027 & 1.045 & 0.310 \\
&  8     & \tb{2.846} & 2.966 & $^*$2.941 & 4.530 & 4.748 & 1.248 
& &\tb{0.956} & $^*$0.996 &1.020 &1.569 &1.737 &0.324\\
&  4    & \tb{4.017} & $^*$4.025 & 4.065 & 5.591 & 6.603 & 1.952 
& &\tb{1.155} & $^*$1.162 &1.204 &1.775 & 1.842 &0.375 \\

\noalign{\global\arrayrulewidth=1pt}
\cline{1-8}\cline{10-15}
\noalign{\global\arrayrulewidth=0.4pt}
\end{tabular}
\label{tab:auc}
}
\end{table*}

\section{Experimental Results}

\subsection{Experimental Setup}

\subsec{Implementation and Training Details}
The overall network architecture and training pipeline follow OmniMVS~\cite{won2020end}, with activation layers in the 3D hourglass network replaced by ELU~\cite{clevert2015fast} and separable 3D convolutions adopted from Birchfield~\etal~\cite{wen2025foundationstereo}, divided into kernels of size $3 \times 3 \times 1$ and $1 \times 1 \times 17$. 
We also construct our own synthetic datasets for training, consisting of two parts: (i) randomly placed objects with backgrounds, and (ii) diverse indoor and outdoor realistic environments. 
The former is used to train the networks from scratch, while the latter is used for fine-tuning. 
We train networks with different numbers of base channels ($C=16, 8,$ and $4$), denoted as $C_{16}$, $C_{8}$, and $C_{4}$, respectively.
The number of inverse-depth indices is set to $D=96$, and the training hyperparameters are fixed to $\lambda_v = 0.5$, $\lambda_c = 1.0$, and $\lambda_n = 1.0$ for all experiments.
%
%
Training is performed on an NVIDIA A100 GPU with a batch size of 4, using SGD with a learning rate of 0.003.

\begin{figure*}[ht]
\centering
    \includegraphics[keepaspectratio=true,width=\textwidth]{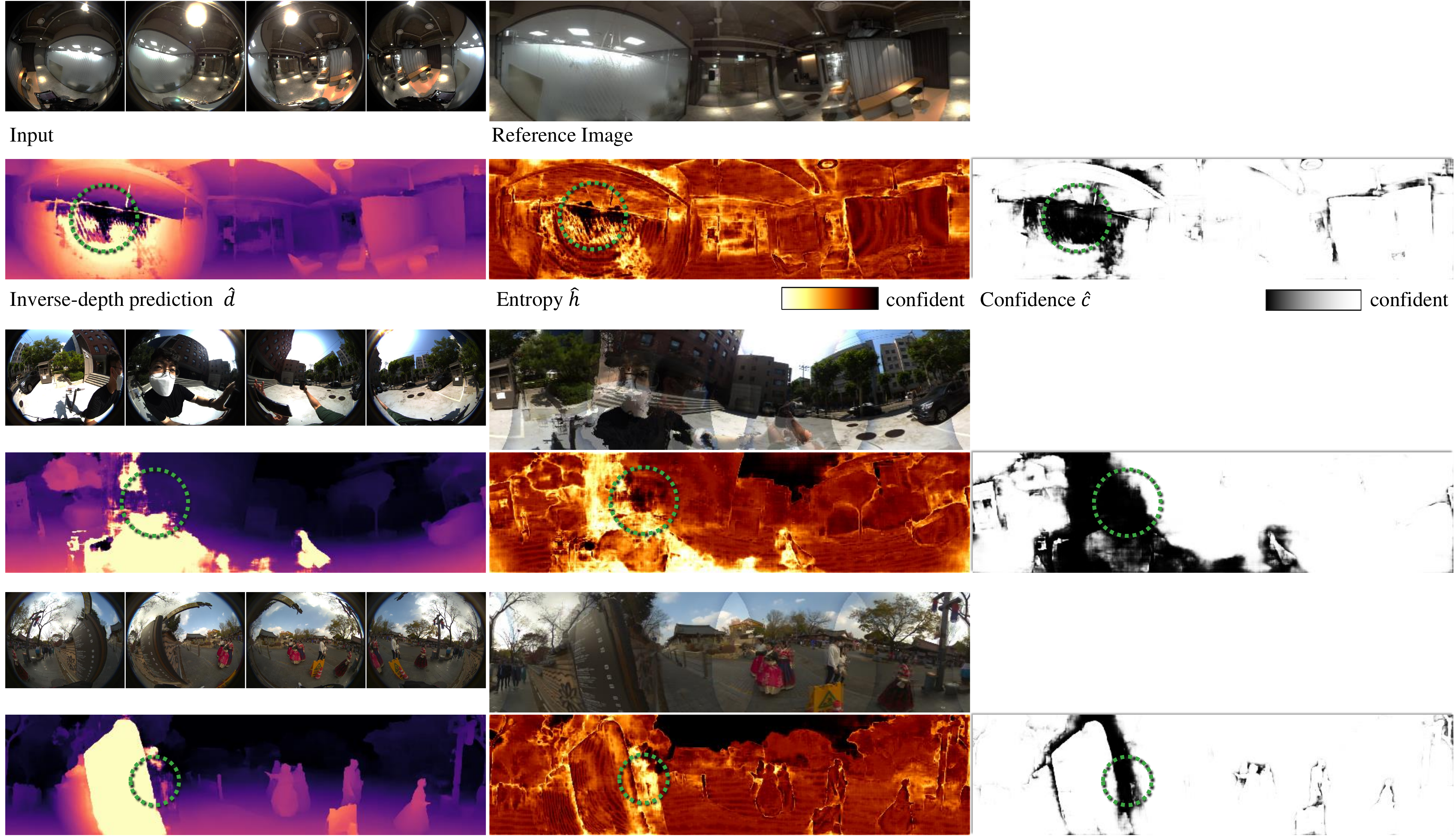}
    \caption{
    \tb{Confidence predictions on real-world scenarios.} 
    Ambiguous regions are indicated by green circles. 
    Note that entropy produces falsely confident pixels in saturated, repetitive, and occluded regions, 
    whereas $\hat{c}$ consistently assigns low confidence values.
    }

    \label{fig:comp_conf}
\end{figure*}


\subsec{Quantitative Evaluation}
We evaluate our method on four synthetic datasets: \emph{Sunny}, \emph{Sunset}, \emph{Cloudy}, and \emph{OmniHouse}~\cite{won2020end}, using ground-truth depth maps. 
%
For \emph{Sunny}, \emph{Sunset}, and \emph{Cloudy}, we evaluate the first 300 frames starting from the first frame respectively, while for \emph{OmniHouse} we evaluate the first 500 frames starting from the first frame.
The percent error of the estimated inverse-depth index is defined as
\begin{equation}
\epsilon = \frac{100}{D} \, \lvert \hat{d} - d^* \rvert.
\label{eq:error}
\end{equation}
For all quantitative evaluations, the output depth map resolution is set to $W=640$ and $H=160$.

\subsec{Qualitative Evaluation}
We also build several capture systems to acquire data across diverse real-world environments, as illustrated in \Figure{fig:introduction}. 
Each system is equipped with four ultra-wide field-of-view (FoV) fisheye cameras mounted on a square-shaped rig with a baseline of $0.2$–$0.3$\,m.
The intrinsic parameters of the lenses and the extrinsic parameters of the rig are calibrated using a checkerboard~\cite{lee2020unified}. 
The rigs are deployed on multiple platforms, including drones, helmet-mounted systems, and wheeled robots, enabling comprehensive evaluation across various real-world scenarios. 
For all qualitative evaluations on real-world data, we use the fine-tuned version of $C_{16}$.

\subsection{Comparison of Confidence Estimation}

In this section, we present both quantitative and qualitative comparisons of the proposed confidence measure against previous approaches: 
the entropy measure~\cite{scharstein1998stereo, won2020end} ($\hat{h}$) and the normalized probability value ($\hat{p}$) at the regressed inverse-depth index~\cite{deng2025omnistereo, li2022mode}, defined as
\begin{equation}
\hat{p} = \mathcal{P}_{\hat{d}}, 
\quad 
\hat{h} = -\sum_{d=0}^{D-1} \mathcal{P}_d \log \mathcal{P}_d.
\label{eq:p_entropy}
\end{equation}
Furthermore, we combine two confidence measures $\hat{c}$, representing the positiveness score, and $\hat{p}$, representing the response normalized along hypotheses, 
to penalize both hard positive and hard negative responses:
\begin{equation}
\gamma_{\alpha,\beta} = (\hat{c})^\alpha (\hat{p})^\beta,
\end{equation}
where $\alpha$ and $\beta$ are sensitivity parameters for each measure. 

To validate the effectiveness of our confidence measure, we compute the area under the curve (AUC)~\cite{gong2005fast,hu2012quantitative,poggi2021confidence}. 
By ranking all assignments in decreasing order of confidence, the AUC of $\hat{c}$ is defined as
\begin{equation}
\text{AUC}(\hat{c}) = 100 \int \frac{|\{x \mid \epsilon_x > \tau,\, x \in M_\rho(\hat{c})\,\}|}{|M_\rho(\hat{c})|}\,d\rho,
\label{eq:auc}
\end{equation}
where $M_\rho(\hat{c}) = \{x \mid \hat{c}_x \ge \operatorname{percentile}_\rho(\hat{c})\}$ and $\tau$ is an outlier threshold. 
In the ideal case where all correct matches are perfectly identified, the theoretical maximum AUC is given by
\begin{equation}
\frac{\text{OptAUC}}{100} = \int_{1-\varepsilon}^{1} \frac{\rho - (1-\varepsilon)}{\rho}\,d\rho
= \varepsilon + (1-\varepsilon)\ln(1-\varepsilon),
\label{eq:opt_auc}
\end{equation}
where $\varepsilon$ is the outlier ratio.
For all evaluations, the outlier threshold is set to $\tau = 1$.
\Table{tab:auc} reports the AUC evaluation of each confidence measure. 
Our confidence measure achieves better performance than both the normalized probability and entropy baselines, 
while the combined confidence further outperforms all other measures, particularly on the \emph{OmniHouse} dataset, 
which contains more textureless and repetitive backgrounds.
This indicates that our approach effectively detects both hard positives and hard negatives.

\Figure{fig:comp_conf} shows a qualitative comparison between our confidence measure $\hat{c}$ and entropy $\hat{h}$, 
obtained by the $C_{16}$ network trained with both $\mathcal{L}_\text{conf}$ and $\mathcal{L}_\text{normal}$ on real-world data. 
%
%
Our proposed confidence measure successfully highlights ambiguous matches. 
In particular, it consistently assigns low confidence to occlusions, 
whereas entropy often yields a few falsely confident outputs.
\begin{table}[t!]
\setlength{\tabcolsep}{2.5pt}         
\caption{
\tb{Quantitative impact of guidance on inverse-depth index errors.} 
Each network is evaluated without fine-tuning on realistic environments. 
The error is defined in \Equation{eq:error}. 
The qualifier \textquotesingle MAE\textquotesingle~denotes the mean absolute error, 
and \textquotesingle RMS\textquotesingle~denotes the root mean squared error. 
All errors are averaged over the test frames, and the lowest value is highlighted.
}

\centering
\resizebox{\linewidth}{!}
{%
\begin{tabular}{cl|rr|rr|rr|rr|rr} \bottomrule
\multicolumn{2}{l|}{Dataset} 
& \multicolumn{2}{c|}{Sunny} 
& \multicolumn{2}{c|}{Cloudy}  
& \multicolumn{2}{c|}{Sunset} 
& \multicolumn{2}{c|}{OmniHouse}  
& \multicolumn{2}{c}{Overall} \\    

\multicolumn{2}{l|}{Metric} 
& \multicolumn{1}{c}{MAE} & \multicolumn{1}{c|}{RMS} 
& \multicolumn{1}{c}{MAE} & \multicolumn{1}{c|}{RMS} 
& \multicolumn{1}{c}{MAE} & \multicolumn{1}{c|}{RMS} 
& \multicolumn{1}{c}{MAE} & \multicolumn{1}{c|}{RMS} 
& \multicolumn{1}{c}{MAE} & \multicolumn{1}{c}{RMS} \\ 
\toprule \bottomrule

\multirow{3}{*}{\rotatebox[origin=c]{90}{$C_{16}$}}
& $\mathcal{L}_{\text{ent}}$~\cite{won2020end} &
1.53 & 6.85  & 2.36 & 10.95 & 1.85 & 8.74 & 3.70 & 12.11 &
2.55 & 10.22 \\ 
& $\mathcal{L}_{\text{conf}}$ &
\tb{1.39} & \tb{6.10}  & 2.04 & 9.30 & \tb{1.31} & 5.89 & 3.41 & 11.32 &
2.23 & 8.93 \\ 
& $\mathcal{L}_{\text{conf}} + \mathcal{L}_{\text{normal}}$ &
1.47 & 6.13 & \tb{1.84} & \tb{7.99} & 1.32 & \tb{5.46} & \tb{2.29} & \tb{8.70} &
\tb{1.81} & \tb{7.43} \\ 
\hline

\multirow{3}{*}{\rotatebox[origin=c]{90}{$C_{8}$}} 
& $\mathcal{L}_{\text{ent}}$~\cite{won2020end} &
2.68 & 11.92 & 3.29 & 14.05 & 2.90 & 12.80 & 6.66 & 19.23 &
4.28 & 15.49 \\
&$ \mathcal{L}_{\text{conf}}$ &
1.93 & 8.05 & 2.06 & 8.48 & 1.89 & 7.65 & 7.25 & 18.86 &
3.85 & 12.99 \\
& $\mathcal{L}_{\text{conf}} + \mathcal{L}_{\text{normal}}$ &
\tb{1.62} & \tb{7.13} & \tb{1.68} & \tb{7.43}  & \tb{1.57} & \tb{7.18} & \tb{3.78} & \tb{12.48} &
\tb{2.39} & \tb{9.46} \\ 
\hline

\multirow{3}{*}{\rotatebox[origin=c]{90}{$C_{4}$}} 
& $\mathcal{L}_{\text{ent}}$~\cite{won2020end} &
3.27 & 12.08 &  3.13 & 11.14 & 4.20 & 14.71 & 9.40 & 23.25 &
5.63 & 17.24 \\
&$ \mathcal{L}_{\text{conf}}$ &
1.98 & 8.18 & 2.68 & 10.43 & 2.06 & 7.51  & 8.67 & 23.29 &
4.54 & 15.50 \\ 
& $\mathcal{L}_{\text{conf}} + \mathcal{L}_{\text{normal}}$ &
\tb{1.40} & \tb{5.25} & \tb{1.62} & \tb{6.70} & \tb{1.45} & \tb{5.74}  & \tb{6.68} & \tb{17.83} &
\tb{3.34} & \tb{11.66}\\ 

\toprule
\end{tabular}
}
\label{tab:depth_metric}
\end{table}

\subsec{Effect of Guidance Change}
We further validate our guidance for confidence prediction by comparing errors without filtering against the previous training strategy based on entropy guidance~\cite{won2020end}:
\begin{equation}
    \mathcal{L}_{\text{ent}} = \mathcal{L}_{\text{smooth-L1}}(\hat{d}, d^*) 
    + \lambda_h \max(\hat{h} - \log{k}, 0),
\end{equation}
where $\lambda_h$ is the weighting parameter of the entropy boundary loss and $k$ is the maximum number of confident indices, set to $1$ and $10$, respectively, during training. 
As shown in \Table{tab:depth_metric}, networks trained with our confidence guidance achieve comparable performance to those trained with entropy guidance, 
demonstrating that the proposed strategy can be applied without adverse effects on depth accuracy.

\begin{figure*}[ht]
\centering
    \includegraphics[keepaspectratio=true,width=\textwidth]{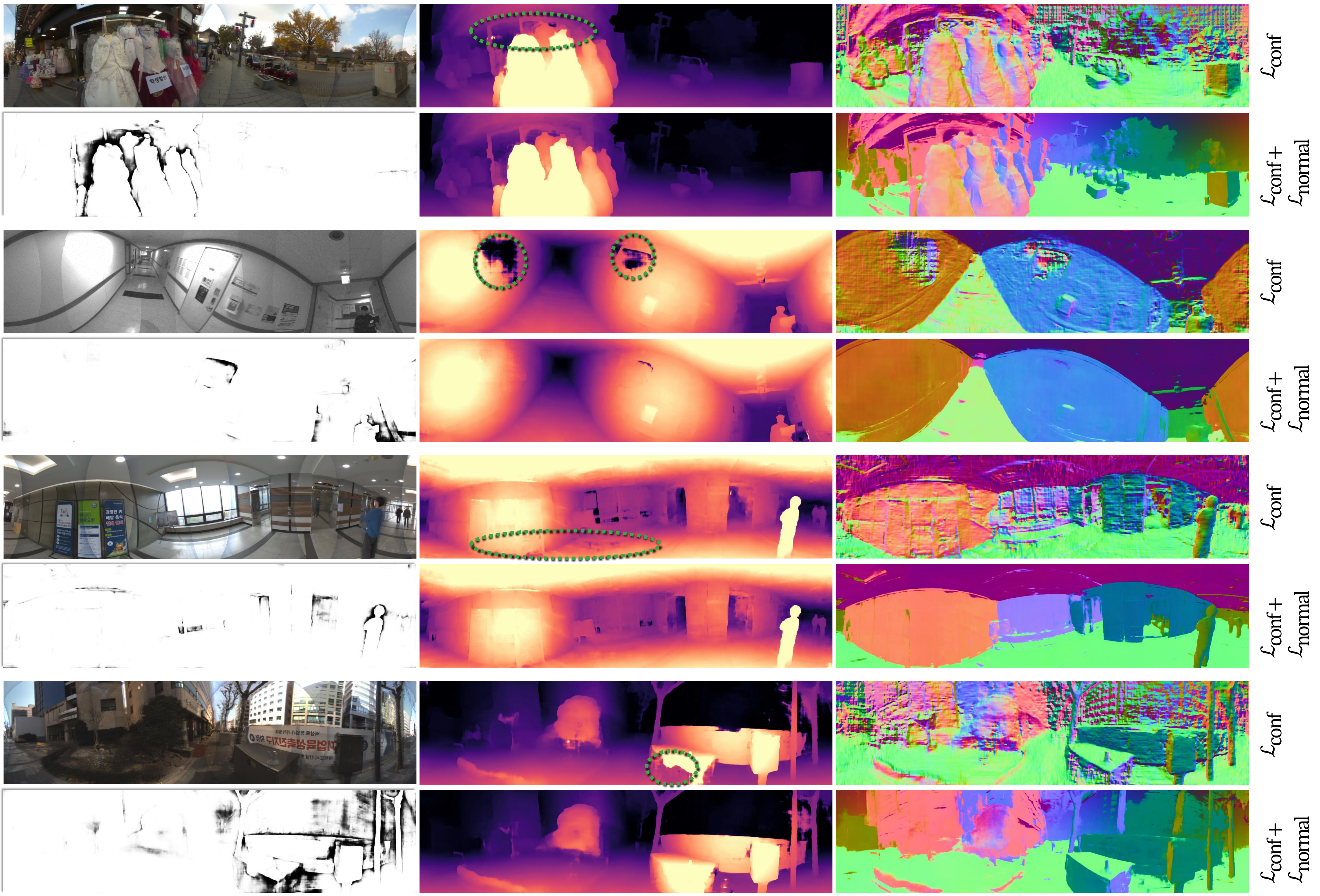}
    \caption{\tb{Qualitative results on real-world scenarios.} 
Erroneous predictions are highlighted by green circles. 
Our proposed method successfully estimates omnidirectional depth, confidence, and surface normals, 
whereas the network without surface normal prediction exhibits errors. 
The loss functions $\mathcal{L}_\text{conf}$ and $\mathcal{L}_\text{normal}$ are defined in \Equation{eq:loss_conf} and \Equation{eq:loss_normal}, respectively.}

    \label{fig:comp_normal}
\end{figure*}

\subsection{Benefits of Joint Surface Normal Prediction}
In this section, we demonstrate the effectiveness of our joint surface normal prediction using the proposed swept feature resampling. 
For the ablation study, we evaluate inverse-depth index errors obtained from networks trained with and without $\mathcal{L}_\text{normal}$ on the synthetic datasets. 
\Table{tab:depth_metric} shows that the network with the surface normal decoder performs favorably compared to the one without it. 
Furthermore, we provide a qualitative comparison on real-world indoor and outdoor environments in \Figure{fig:comp_normal}. 
For the network without a surface normal prediction head, normal maps are computed by estimating the plane normal of 3D points within an $11 \times 11$ window in the output inverse-depth map. 
The network trained with joint normal prediction successfully compensates for erroneous estimates in saturated, textureless, and reflective regions as well as along object boundaries.
This demonstrates that joint learning with swept feature volume resampling enables the response-volume aggregation stage to incorporate contextual cues more effectively.

\section{CONCLUSIONS}
In this paper, we proposed a training strategy for confidence estimation in omnidirectional stereo to mitigate ambiguous matches that frequently occur in wide-baseline multi-camera rigs. 
We also introduced \emph{Swept Feature Volume Resampling}, which addresses the challenge of exploiting input meta-information caused by the difference between the optical centers of the inputs and the output. 
The proposed confidence estimation requires no additional network modules, and \emph{Swept Feature Volume Resampling} enables joint learning of surface normals. 
Extensive experiments on both real and synthetic datasets demonstrate that our approach more effectively suppresses ambiguous regions compared to prior confidence estimation methods. 
Moreover, surface normals learned via \emph{Swept Feature Volume Resampling} improve depth predictions in real-world indoor scenes, highlighting the advantage of jointly leveraging meta-information for robust omnidirectional depth estimation.
Finally, since our method introduces minimal architectural overhead and avoids auxiliary inference stages, it remains practical for real-time deployment on mobile robotic platforms such as drones, wheeled robots, and wearable systems.

\bibliographystyle{IEEEtran}
\bibliography{reference}

\end{document}